\def\year{2022}\relax
\documentclass[letterpaper]{article} % DO NOT CHANGE THIS
\usepackage{aaai22}  % DO NOT CHANGE THIS
\usepackage{times}  % DO NOT CHANGE THIS
\usepackage{helvet} % DO NOT CHANGE THIS
\usepackage{courier}  % DO NOT CHANGE THIS
\usepackage{url}  % DO NOT CHANGE THIS
\usepackage{todonotes} % to add notes
\usepackage{graphicx} % DO NOT CHANGE THIS
\usepackage{graphicx}  % DO NOT CHANGE THIS
\usepackage{array}
\usepackage{verbatim}
\usepackage{svg}
\usepackage{amsmath}
\usepackage{multirow}
\usepackage{subcaption}
\usepackage{csquotes}
\usepackage{natbib}

\usepackage[many]{tcolorbox}
\usepackage{xcolor}
\usepackage{varwidth}
\usepackage{environ}
\usepackage{xparse}
\usepackage{makecell}
\usepackage[normalem]{ulem}
\usepackage{bm}

\title{Linguistic Characterization of Divisive Topics Online: Case Studies on Contentiousness in Abortion, Climate Change, and Gun Control} % TBD - Diyi
 \author{
    Jacob Beel, \textsuperscript{\rm 1} Tong Xiang, \textsuperscript{\rm 2} Sandeep Soni, \textsuperscript{\rm 1} Diyi Yang \textsuperscript{\rm 1}
    
 }
 
 \affiliations{
    \textsuperscript{\rm 1} Georgia Institute of Technology\\%, School of Interactive Computing, Atlanta, Georgia 30318 \\
    \textsuperscript{\rm 2} Georgetown University\\%, Department of Computer Science, Washington, DC 20057 \\
    jbeel3@gatech.edu, tx39@georgetown.edu, sandeepsoni@gatech.edu, dyang888@gatech.edu
 }

\newcommand{\Rq}[1]{\textbf{RQ#1}}

\begin{document}

% \nocite{*}

\maketitle

\begin{abstract}

As public discourse continues to move and grow online, conversations about divisive topics on social media plat-forms have also increased. These divisive topics prompt both contentious and non-contentious conversations. Although what distinguishes these conversations, often framed as what makes these conversations contentious, is known in broad strokes, much less is known about the linguistic signature of these conversations. Prior work has shown that contentious content and structure can be a predictor for this task, however,most of them have been focused on conversation in general,very specific events, or complex structural analysis. Additionally, many models used in prior work have lacked interpretability, a key factor in online moderation. Our work fills these gaps by focusing on conversations from highly divisive topics (abortion, climate change, and gun control), operationalizing a set of novel linguistic and conversational characteristics and user factors, and incorporating them to build interpretable models. We demonstrate that such characteristics can largely improve the performance of prediction on this task, and also enable nuanced interpretability. Our case studies on these three contentious topics suggest that certain generic linguistic characteristics are highly correlated with contentiousness in conversations while others demonstrate significant contextual influences on specific divisive topics.
\end{abstract}
\section{Introduction}
Divisive topics -- from gun control to illegal immigration, from animal cruelty to vaccination in children -- have gained another avenue for exposure in online media.
These topics evoke opinionated and polarizing conversations and though we broadly understand the different issues that make these topics contentious, we know very little about the structure and dynamics at the level of conversations around these topics. For instance, what makes a conversation about a particular topic contentious? What are the differences between conversations which foster healthy disagreement and discussion about a divisive topic to those which devolve into online shouting matches? What is the extent of this effect in conversations about a particular topic on a particular platform?

Answers to the above questions and a nuanced understanding of these conversations could help in resource allocation on social media platforms: for example, determining the course of a conversation early on in its lifespan could help to guide moderator action and perform a kind of triage for online community moderation.
Additionally, the distribution of these conversations on the platform can help in user enculturation: for instance, platforms can guide newcomers so that they are exposed to less contentious conversations and get a gentle introduction to the communities dedicated to discussing divisive topics. Alternatively, measuring the extent of contentious conversations can be used to assess platform health \citep{matamoros2017platformed}: for example, conversations that manifest as constant inter-group conflict or lead to hateful speech or vile ad hominem attacks can be taken as a bad sign, marking platforms as amplifiers and manufacturers of negative discourse; in contrast, discussions that balance between opposing viewpoints or improve information diversity can be viewed as good signals. 

Much attention has recently been paid towards predicting conversation quality, typically framed as controversy, from the content of and structure of conversations~\cite{hessel2019something,kim-2014-convolutional}. The strength of these works lies in the use of powerful neural networks, such as convolutional neural networks (CNN) and transformers (BERT), to predict controversy in conversations with high accuracy. However, these predictive models are constrained by their relative limitations in highlighting the informative conversation characteristics that distinguish controversial conversations from non-controversial ones. Moreover, their predictive performance has only been tested in generic conversations, making it hard to tease out conversation topics being divisive from participation patterns in discussions leading to controversy. This work fills this gap by focusing exclusively on divisive topics.
In this work, we aim to address the above gaps by answering the following research questions:
\Rq{1} Can we accurately predict contentiousness in conversations via explainable 
conversation characteristics when controlling for divisive topics? \Rq{2} What specific conversation characteristics differentiate  non-contentious conversations from contentious ones between certain divisive topics? 

To address RQ1 we
operationalize a set of linguistically explainable characteristics including coarse discourse acts, toxicity, sentiment, and occurrence of specific lexicons. In addition, we propose to leverage conversation participants' characteristics such as their gender, location, account age, and prolificity. Many of these features, including discourse acts and user features have not been used in this task previously. 
We follow prior work~\citep{choi2010identifying,hessel2019something}
to define quality of conversations. 
We focus on conversations identified to belong of one of the three divisive topics: abortion, climate change, and gun control and then look to examine the contentiousness of conversations within those divisive topics.
These conversation characteristics are then utilized to build machine learning classifiers to predict whether a conversation is contentious. 
To address RQ2, 
we visualize 
the most predictive 
conversation characteristics, 
across and within each divisive topic. 
Our analyses show that certain generic linguistic characteristics are highly correlated with contentiousness in conversations across the board: for instance, negative discourse acts appear significantly more often in contentious conversations across three topics; while others demonstrate significant contextual influences such as the varying importance of user demographic information.

To sum up, the contributions of our work are three-fold. First, we 
operationalize a set of conversation characteristics such as coarse discourse acts and toxicity, which can improve both the predictive performance and the explainability of contentious conversation detection. Second, 
we demonstrate that the contentiousness can be detected using our explainable models with reasonable accuracy, even at an early stage of these conversations. 
Third, our research uncovers both generic and topic specific conversation characteristics associated with discussions about divisive topics online. This work can be leveraged for several different purposes, including further research and aiding moderation strategies of online communities.

Note that this work uses terminology which may be different from prior work due to the structure of this research. For instance, \citeauthor{hessel2019something} refer to conversations as ``controversial''. In this work, we elect to refer to conversations as ``contentious'' and to topics as ``divisive'' in order to ensure that the difference between the nature of the two is as clear as possible. The topics we study are divisive, but will not necessarily result in contentious conversations.

\section{Related Work}
\subsection{Controversy in Online Discussions}
There has been extensive computational research about controversy in online discussions. 
To quantify controversy, \citeauthor{choi2010identifying} (2010) 
used the ratio of positive and negative sentiment words; 
\citeauthor{rethmeier-etal-2018-learning} (\citeyear{rethmeier-etal-2018-learning}) and \citeauthor{hessel2019something} (\citeyear{hessel2019something}) define controversial content as that which attracts a mix of positive and negative feedback. 
Other lines of research used predefined lexicons to collect controversial discussions \cite{awadallah2012harmony}.
Prior work on predicting controversy or distinguishing controversial discussions from non-controversial ones rely heavily on dictionary based approaches~\cite{pennacchiotti2010detecting,addawood-etal-2017-telling}. Such methods benefit from being explainable but have limited predictive power \cite{dori2013detecting}. 
Supervised neural models such as CNNs have also been used recently for predicting controversy~\cite{rethmeier-etal-2018-learning}. Since it is difficult to get ground-truth annotations of controversy, these approaches typically use distant supervision, such as upvote-downvote ratio, to mark examples as either controversial or non-controversial.
The growing popularity of Reddit has attracted research attention to predict and model controversy on its platform \cite{conover2011political}. \citeauthor{hessel2019something} (2019)  uses BERT to predict controversial conversations on Reddit.  \citeauthor{guimaraes2019analyzing} (2019) refines controversy into disputes, disruptions, and discrepancies via user actions and connects those user actions to sentiment and topic analysis to explain controversy specifically within political discussion. 
\subsection{Contentiousness, Divisive Topics and Controversy}
Different from all the aforementioned research, we argue that controversy should be studied at least from two perspectives: topics being divisive, and conversations being contentious.
The key intuition is that, people can still have deliberative and good discussions regarding a very divisive topic such as gun control.
Thus, topics being divisive does not equal to conversations being contentious, and analyzing how conversation characteristics relate to contentious content should control for divisive topics. 
For the purposes of our work, controversial and non-controversial conversations are referred to as contentious and non-contentious, respectively. This is more accurate to our task at hand, which is to distinguish between two types of discourse which emerge when even discussing divisive topics rather than detect controversy irrespective of topic. Therefore, 
our work controls for topics in order to explore why certain conversations become controversial and why others do not within and between those topics. We do this by opting to focus on linguistic characterizations of conversations as well as user factors rather than conversation structure and comment timing. In contrast to both \citeauthor{rethmeier-etal-2018-learning} and \citeauthor{hessel2019something}, we focus on using linguistic characteristics and user factors to build simpler models, rather than end-to-end neural solutions. In doing so we create a more explainable solution, and compare these models against the transformer architecture BERT~\cite{devlin2018bert}. 

\subsection{Abortion, Climate Change, and Gun Control}
To better study conversation characterises and their influences on conversations being contentious, we first need to control for divisive topics, one big confounding factor behind controversy. 
As a case study, we focus on discussion around three topics -- abortion, climate change, and gun control. These contentious topics have long been in the public discourse~\cite{schweizer2013changing} but have gained an additional dimension: views on these topics are now recorded permanently on social media platforms making these platforms attractive venues for research on the social aspects of conversations~\cite{tremayne2013opinion,wasike2017persuasion,chor2019factors}.
Online discussions about these contentious topics tends to be polarizing~\cite{garimella2017ebb}. For instance, \citet{garimella2017effect} found that debates on contentious topics, such as gun control or abortion, on Twitter become more polarizing as a result of increased public attention that typically follows some external event. Similarly, \citet{demszky2019analyzing} measured polarization based on the content of tweets around issues of gun policies for a series of documented mass shooting events. However, in contrast to these objectives of measuring polarization around these topics, our focus is on characterizing what linguistic features are correlated with controversy. 
The topic of climate change, though relatively recent compared to gun control and abortion, has also attracted research particularly on how it is discussed online~\cite{olteanu2015comparing,anderson2017social}. 
\citet{jang2018explaining} proposed a method to summarize discussions on contentious topics such as abortion and climate change, to better understand the controversy and these different viewpoints. Our focus in this work is also on explaining controversy in these topics but we diverge in two ways: first, we focus on understanding the linguistic aspects that are predictive of controversy in conversations; second, we use high-capacity neural networks to classify conversations as controversial or non-controversial, allowing us to distinguish between the two for any of the contentious topics that we considered. 
\section{Data}
\label{sec:data}
Our data consist of filtered and labeled posts and comments from Reddit.  
Reddit is a content-sharing and discussion social media website where users can post conversations, which other users can comment on. The data consist of most conversations (also known as ``posts'' or ``threads'') and comments over the period of time from 2007 to February 2014~\cite{hessel2019something}, and are drawn from the same source as is used by~\citeauthor{hessel2019something}. Deleted conversations are not included, but conversations and comments which were made by accounts that have since been deleted are included. A conversation consists of a title, selftext (the body of an original post), and comments. The title and selftext are input by the original poster (OP) at the time of conversation creation. We label the posts according to the scheme proposed and validated in~\citeauthor{hessel2019something} (\citeyear{hessel2019something}). Under this scheme, a conversation is considered contentious if its upvote ratio is within the bottom quartile of conversations, i.e. it has drawn more mixed user reactions. Only conversations which have an upvote ratio of at least $0.5$ are considered.

Although we use the same labelling technique as \citeauthor{hessel2019something}, we control for divisive topics. This allows us explore the manner in which contentious conversations differ within those divisive topics, a question which~\citeauthor{hessel2019something} elect not to explore. Additionally, we explore a larger number of subreddits by virtue of our choice to filter by topic. As a robustness check, we also compared this labelling technique to another which is widely used in this task \cite{choi2010identifying}, the ratio of positive to negative sentiment. Across all topics, this resulted in unbalanced data sets, with non-contentious conversations making up over $85\%$ of conversations. We found this definition of controversy to differ significantly, and state of the art models can not perform significantly better than random chance. These two labelling techniques clearly measure different phenomena, and we urge future work to explore these differences for deeper insights.

To retain only the conversations associated with particularly divisive topics, %When selecting the topics we were to study, 
we began with a larger list of divisive topics available at ProCon.org (\url{http://procon.org}) and filtered down to a subset of topics that met our selection criteria. First we aimed to maximize the number of conversations. Second, we aimed to minimize the number of off-topic conversations through a process of filtering, followed by manual inspection, followed by refinement of key words. This process was repeated until the number of off-topic conversations was minimized. Finally, we aimed for topics with a large difference between the upvote ratios of the two label categories. Topics which did not satisfy this criteria, such as the ``\emph{U.S. involvement in the Middle East}'', ``\emph{free or affordable higher education}'', ``\emph{healthcare}'', and ``\emph{police brutality}'', were excluded. The topics we focus on and the number of conversations $n$ in each topic are: abortion (n=892), climate change (n=1070), and gun control (n=1176). These were selected due to their large differences aside from being political and divisive.

These discussions (conversations and comments) were identified using keywords selected from the ProCon.org pages for these topics: abortion (e.g., \emph{``pro-life'', ``pro-choice'', ``fetus'', ``abortion''}), climate change (e.g., \emph{``global warming'', ``climate change'', ``greenhouse gas'', ``fossil fuel''}), and gun control (e.g., \emph{ ``ar-15'', ``national rifle association'', ``assault rifle'', ``second amendment''}). The three selected topics had the highest number of conversations of the topics we considered and a wide enough range of upvote ratio to indicate the presence of different classes of conversations. These topics also minimized the number of false positives, conversations that matched one or more keywords but did not actually pertain to the topic. Additionally, these topics presented ways in which they may differ. For instance, we hypothesized that abortion would be a topic where a non-minority presence of male users in a conversation would lead to contentiousness, whereas a non-minority presence of female users in a conversation would not.

\section{Conversation Characteristics} \label{sec:features}
We operationalize a set of conversation characteristics that include both linguistic and non-linguistic features needed by subsequent models and explainable analyses. 

\subsection{Linguistic Operationalization}
\textbf{Discourse Acts and Discussion Patterns}
contentiousness in a discussion is reflected by the expression of opposing views or disagreements. To better understand those discourse patterns in contentious conversations, we leverage 
coarse discourse acts \cite{zhang2017characterizing} 
to extract different types of patterns in users' discussions. 
These discourse acts consist of the following 10 acts: \textit{question}, \textit{answer}, \textit{announcement}, \textit{agree}, \textit{appreciate}, \textit{disagree}, \textit{negative}, \textit{elaborate}, \textit{humor}, and \textit{other}. 
We used BERT and fine-tuned it to predict these discourse acts for each post and each comment using the Reddit annotated data from  \citeauthor{zhang2017characterizing} (\citeyear{zhang2017characterizing}), with $65\%$ 10-class macro accuracy at predicting these discourse acts. 
The entire conversation conversation can then be represented in terms of these discourse acts as unigram and bigram vectors. \smallskip

\noindent
\textbf{Toxicity} 
Previous work has shown that toxicity is linked with contentious conversations on social media~\citep{zhang-etal-2018-conversations,mejova2014controversy}. Prior work has leveraged it as a signal for unhealthy interactions in online conversations~\citep{raman2020stress}. We design toxicity features in order to evaluate the rudeness, disrespectfulness, or unreasonableness of posts within a conversation. We use Google's Perspective API\footnote{\url{https://www.perspectiveapi.com/}} to extract the toxicity of text. The API returns a value between $0$ and $1$ for a given input, where $0$ indicates complete inoffensiveness and $1$ indicates absolute offensiveness. 
We construct three features with the aforementioned toxicity values: \textit{title toxicity} to represent the toxicity of the title, \textit{post toxicity} to represent the toxicity of the post, and \textit{max. comment toxicity} to represent the most toxic comments in the conversation conversation.\smallskip

\noindent
\textbf{Sentiment}
Sentiment is also highly related to contentiousness~\cite{mishne2006leave,pennacchiotti2010detecting,mejova2014controversy}. We use Valence Aware Dictionary and sEntiment Reasoner (VADER)~\cite{ICWSM148109} to measure the sentiment of posts and their comments. 
We use the built-in VADER lexicons from \texttt{NLTK}~\cite{bird-loper-2004-nltk} to obtain a sentiment score for any given text. 
We measure the level of positivity and negativity in the post (\textit{max. sentiment}, \textit{min. post sentiment}),  
the maximum level of positivity in the comments (\textit{max. comment sentiment}), 
the average amount of positivity in the comments (\textit{avg. comment positive sentiment}), 
as well as  the maximum and average amount of negativity in the comments (\textit{min. comment sentiment}, and \textit{avg. comment negative sentiment}) and use them as separate features for prediction. \smallskip 

\noindent
\textbf{LIWC}
The linguistic styles and word choices that speakers choose to convey their intent can reflect how they conceptualize meaning~\citep{trott-etal-2020-construing}. Prior studies suggested that information gained from examining these lexical patterns can be useful in contentiousness prediction~\citep{addawood-etal-2017-telling,addawood-bashir-2016-evidence}. To understand how linguistic styles affect contentious conversations, we employ the Linguistic Inquiry and Word Count (LIWC)~\citep{pennebaker2001linguistic,pennebaker2007development}. We construct a set of features to help capture lexical evidence using its $64$ dimensions, including aspects such as attentional focus, social relationships, thinking styles, and others~\citep{tausczik2010psychological},
by extracting a count vector of occurrences of words from the LIWC lexicon. \smallskip

\noindent
\textbf{Text Representations}
We choose to represent the meaning of the text of each conversation conversation in two vector representations: TF-IDF and contextual representation from a finetuned BERT model. The TF-IDF representation is the sum of the tf-idf vector for the post and the average tf-idf vector of all the comments under the post. We set a maximum document frequency of $0.95$ and exclude English stopwords. The BERT representation is computed using bert-base-uncased from \texttt{transformer}~\cite{wolf2019transformers} and is similarly the sum of the contextual representation for the post and the average comment representation. We use the second-to-last hidden states as the representation vectors. 

\subsection{User Factors}
We also include a set of user factors that are closely connected with contentiousness \cite{addawood-etal-2017-telling}, which are often leveraged to understand the types of people who are participating in these conversations. 
We use the gender, location, and prolificity of Reddit users as operational features. 
Our choice of restricting to these factors is partially because this information can be extracted more easily and with high confidence from Reddit. Additionally, gender is often associated with different roles in online conversations and reactions to contentiousness \cite{herring2020dominates}. Similarly, location is informative due to its contextualized effect in understanding contentiousness in different communities \cite{chen2013and}. Prolificity can be used to distinguish between newcomers and experienced members.

Gender and location are heuristically determined by textually examining a user's post history for noun phrases. Gender is determined by looking for gendered nouns or pronouns at the end of a noun phrase taking the form ``I am...'', or any other form of the copula. Genders are counted, with indeterminate and nonbinary genders being excluded from this count. Locations are extracted in a similar fashion, and vectorized as a count vector of geocoded locations or the raw location if the location fails to be geocoded. The distribution of user locations differs from the distribution of user locations generally, indicating a connection with contentiousness. Prolificity is calculated from the number of posts a user has made in conversations on the topic of interest. We define a user as prolific if she/he appears in her/his respective topic more than $25$, $50$, or $100$ times in order to explore differences between different levels of prolificity.

To ensure quality, we validated the gender and locations of users extracted using the above steps.
We find a female/male split of $47.84\%$/$52.16\%$, $25.56\%$/$74.44\%$, and $33.32\%$/$66.68\%$ for abortion, climate change, and gun control, respectively.
The overall gender distribution closely matches surveyed Reddit demographics.\footnote{\small{https://www.journalism.org/2016/02/25/reddit-news-users-more-likely-to-be-male-young-and-digital-in-their-news-preferences}} We also use location to validate the corpus we collected, where we find that the U.S. is the most common country at $44\%$, followed by Canada, India, the U.K, Brazil, and Germany. This does not perfectly mirror reported demographics, but comes close.\footnote{https://www.statista.com/statistics/325144/reddit-global-active-user-distribution}

\subsection{Subreddit}

The subreddit of a conversation is included as a one-hot encoded feature to account for its potentially confounding influence. We found no statistically significant ($p < 0.05$) evidence that these distributions were different using the Kolmogorov-Smirnov 2 sample test \cite{smirnov1939estimation}. The $p$ for abortion, climate change, and gun control are $0.88$, $0.60$, and $0.80$ respectively.

\section{Prediction of Contentious Conversations}
\label{sec:prediction}

\begin{table*}[ht!]
\setlength{\tabcolsep}{3pt}
\renewcommand{\arraystretch}{1}
\small
\begin{center}
\begin{tabular}{|l|l|l|l|l|l|l|l|l|l|l|l|l|}
\hline
\multirow{3}{*}{Features} & \multicolumn{4}{c|}{Abortion} & \multicolumn{4}{c|}{Climate change} & \multicolumn{4}{c|}{Gun control} \\ \cline{2-13}
& Acc.    & Prec.    & Rec.   & F1   & Acc.   & Prec.   & Rec.   & F1  & Acc.  & Prec.  & Rec.  & F1  \\
\hline
\hline
TF-IDF
& $0.586$ & $0.578$ & $0.583$ & $0.546$
& $0.644$ & $0.614$ & $0.581$ & $0.581$
& $0.653$ & $0.640$ & $0.617$ & $0.613$\\\hline
+ Discourse act
& $0.660^{*}$ & $0.628^{*}$ & $0.613$ & $0.599$
& \bm{$0.696$} & $0.649$ & $0.617$ & $0.624$ 
& \bm{$0.721^{*}$} & \bm{$0.694^{*}$} & $0.657$ & $0.668$\\\hline
+ Gender
& $0.644^{*}$ & $0.620^{*}$ & $0.595$ & $0.585$
& $0.639$ & $0.611$ & $0.569$ & $0.575$
& $0.685$ & $0.667^{*}$ & $0.614$ & $0.630$\\\hline
+ LIWC
& $0.665^{*}$ & $0.650$ & $0.617$ & $0.608$
& $0.672$ & \bm{$0.671^{*}$} & $0.554$ & $0.590$
& $0.687$ & $0.685$ & $0.660$ & $0.652$\\\hline
+ Location
& $0.624$ & $0.594$ & $0.581$ & $0.567$
& $0.641$ & $0.617$ & $0.586$ & $0.582$
& $0.693$ & $0.672$ & $0.644$ & $0.647$\\\hline
+ Prolific
& $0.657^{*}$ & $0.618^{*}$ & $0.606$ & $0.596$
& $0.669^{*}$ & $0.640^{*}$ & $0.605^{*}$ & $0.607^{*}$
& $0.703^{*}$ & $0.684^{*}$ & $0.630$ & $0.648$\\\hline
+ Sentiment
& $0.596$ & $0.581$ & $0.596$ & $0.556$
& $0.676$ & $0.634^{*}$ & $0.625$ & $0.616$
& $0.665$ & $0.650$ & $0.625$ & $0.625$\\\hline
+ Toxicity
& $0.609$ & $0.586$ & $0.608$ & $0.565$
& $0.661$ & $0.638$ & $0.614$ & $0.610$
& $0.660$ & $0.646$ & $0.618$ & $0.618$\\\hline
+ Subreddit 
& $0.550$ & $0.551$ & $0.547$ & $0.548$
& $0.649$ & $0.659$ & $0.607$ & $0.631$
& $0.597$ & $0.610$ & $0.541$ & $0.573$\\\hline
+ All 
& \bm{$0.704^{*}$} & \bm{$0.753^{*}$} & \bm{$0.605$} & \bm{$0.663$} 
& \bm{$0.707$} & \bm{$0.766$} & $0.593$ & \bm{$0.660$}
& $0.683^{*}$ & \bm{$0.727^{*}$} & $0.573$ & $0.631$\\\hline
\hline
BERT
& $0.664$ & $0.652$ & $0.641$ & $0.617$
& $0.757$ & $0.729$ & $0.678$ & $0.693$
& $0.728$ & $0.701$ & $0.709$ & $0.692$\\\hline
+ Discourse
& $0.702$ & $0.664$ & $0.638$ & $0.639$
& $0.772$ & $0.754$ & $0.692$ & $0.716$
& $0.746$ & $\bm{0.726}$ & $0.703$ & $0.707$\\\hline
+ Gender
& $0.660$ & $0.620$ & $0.623$ & $0.605$
& $0.761$ & $0.724$ & $0.701$ & $0.702$
& $0.739$ & $0.706$ & $0.714$ & $0.702$\\\hline
+ LIWC
& $0.686$ & $0.639$ & $0.668$ & $0.635$
& $0.761$ & $0.736$ & $0.686$ & $0.701$
& $0.734$ & $0.705$ & $0.695$ & $0.692$\\\hline
+ Location
& $0.652$ & $0.626$ & $0.699$ & $0.630$
& $0.746$ & $0.712$ & $0.689$ & $0.689$
& $0.741$ & $0.723$ & $0.698$ & $0.701$\\\hline
+ Prolific
& $0.698$ & $0.652$ & $0.688$ & $0.652$
& $\bm{0.780}$ & $0.742$ & $\bm{0.714}$ & $\bm{0.719}$
& $\bm{0.752}$ & $0.724$ & $\bm{0.716}$ & $\bm{0.711}$\\\hline
+ Sentiment
& $0.657$ & $0.653$ & $0.632$ & $0.617$
& $0.768$ & $0.744$ & $0.708^{*}$ & $0.717^{*}$
& $0.732$ & $0.707$ & $0.703$ & $0.693$\\\hline
+ Toxicity
& $0.648$ & $0.612$ & $0.683$ & $0.618$
& $0.760$ & $0.735$ & $0.689$ & $0.700$
& $0.728$ & $0.701$ & $0.698$ & $0.688$\\\hline
+ Subreddit
& $0.558$ & $0.552$ & $0.619$ & $0.582$
& $0.666$ & $0.652$ & $0.708$ & $0.678$
& $0.594$ & $0.580$ & $0.689$ & $0.629$\\\hline
+ All
& $\bm{0.715}$ & $\bm{0.686}$ & $\bm{0.712}$ & $\bm{0.678}$
& $0.774$ & $\bm{0.755}$ & $0.685$ & $0.712$
& $0.740$ & $0.723$ & $0.707$ & $0.706$\\\hline
\end{tabular}
\end{center}
\caption{Comparison of the performance for each feature set. We highlight the model with the best performance in bold. $^*$ denotes having statistically significant differences from their corresponding baselines at the $p<.05$ level.
}
\label{table:Evaluation results}
\setlength{\tabcolsep}{6pt}
\renewcommand{\arraystretch}{1}
\end{table*}

To answer RQ1, we train classifiers to predict whether a conversation is labeled as contentious or otherwise, incorporating the linguistic characteristics and the user factors as features along with subreddit in the classifier.
Specifically, we show the performance of the classifier in making two types of predictions: predicting the contentiousness of the conversation using all the comments and predicting the contentiousness at different stages of these conversations. 
We experimented with two classifiers as baselines: Logistic Regression (LR) using TF-IDF features and Multi-layer Perceptron (MLP) using BERT features. We then augmented these baselines to include the additional linguistic and user characteristics described in Section 4. All the models were implemented using the off-the-shelf machine learning library \texttt{scikit-learn}. The pretrained BERT representation was further finetuned on Reddit comments before being used as input for the MLP. 10-fold stratified cross-validation was used to evaluate all the models and account for potentially confounding factors. The resulting scores are displayed in Table~\ref{table:Evaluation results}.
\subsection{Conversation Level Contentiousness Prediction}
As shown in Table~\ref{table:Evaluation results}, 
augmenting the baselines with the linguistic and the user features boosts the prediction performance significantly across all three topics. Specifically, in all cases, TF-IDF + All outperforms the baseline TF-IDF. Similarly, BERT + All also displays significant improvements over its baseline though not to the same degree as the TF-IDF baseline. A closer look reveals that the addition of certain features improves the results across all topics.
On the other hand, some features only provide topic-specific improvements -- for instance, adding the gender features does not improve the performance for the climate change topic. This suggests a distinction between features that are more related to contentiousness in general, and features that are more related to a specific divisive topic. Our modeling of contentiousness through this combination of linguistic and user features highlights these differences which would otherwise go unnoticed if using just the TF-IDF or BERT representations. 

\begin{figure*}[ht!]
\centering
\begin{subfigure}[b]{0.3\textwidth}
\centering
  \includegraphics[width=\textwidth]{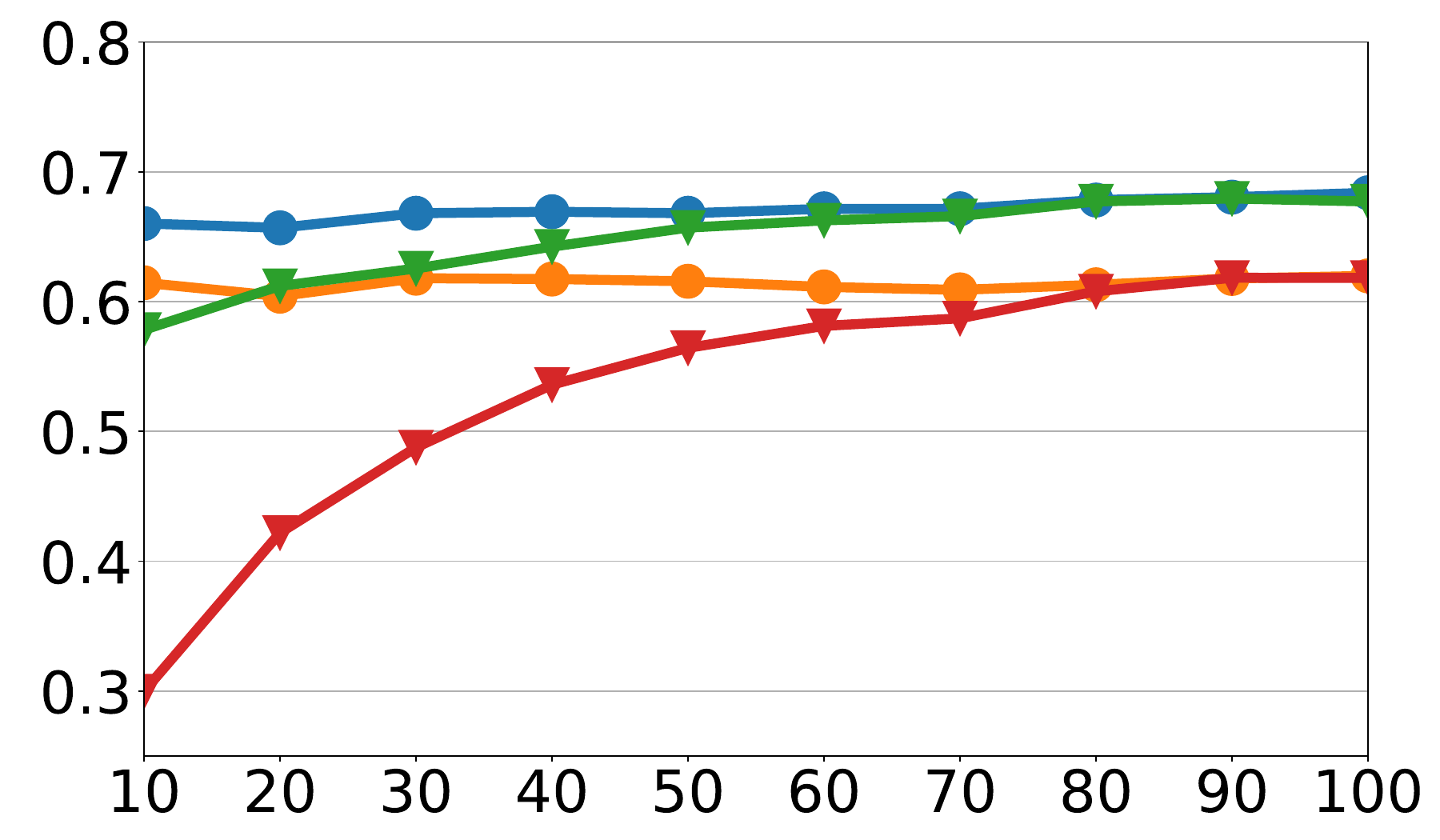}
  \caption{Abortion}
  \label{fig:partial_prediction_abortion}
\end{subfigure}
\hfill
\begin{subfigure}[b]{0.3\textwidth}
\centering
  \includegraphics[width=\textwidth]{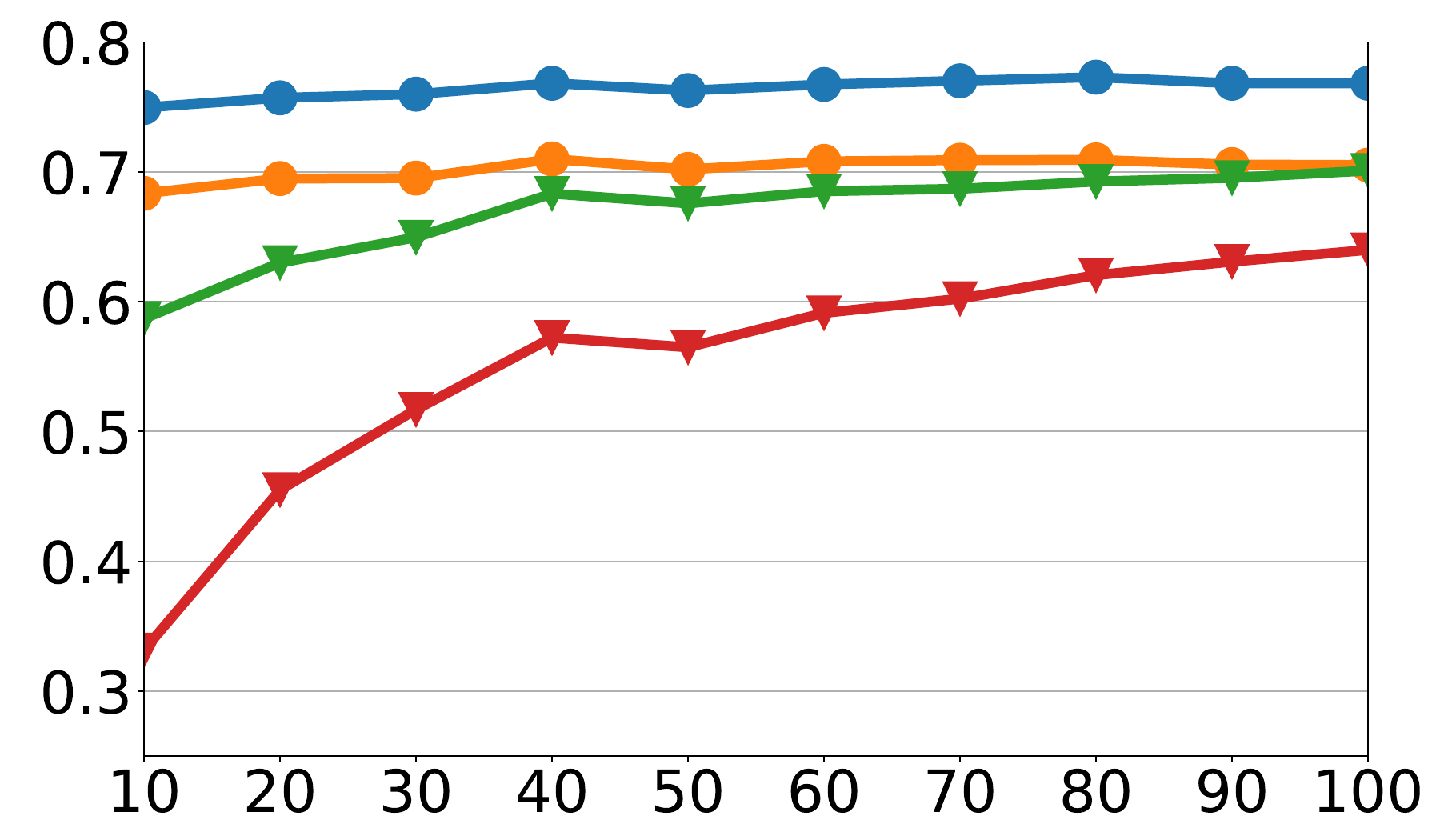}
  \caption{Climate Change}
  \label{fig:partial_prediction_climate_change}
\end{subfigure}
\hfill
\begin{subfigure}[b]{0.3\textwidth}
\centering
  \includegraphics[width=\textwidth]{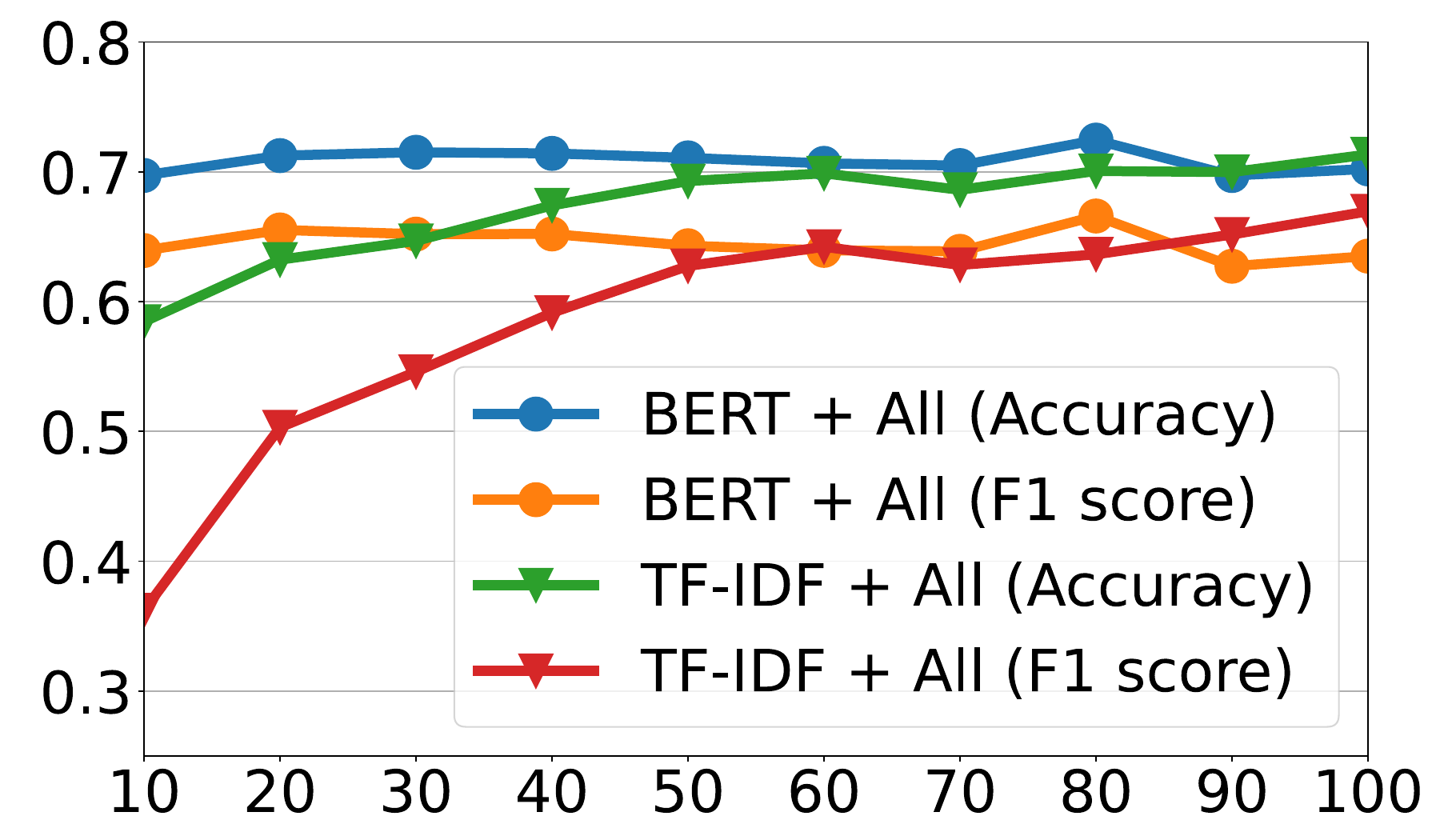}
  \caption{Gun Control}
  \label{fig:partial_prediction_gun_control}
\end{subfigure}
\caption{Accuracy and F1 score for early prediction of contentiousness using TF-IDF + All and BERT + All models. The percentage of comments used as input is shown on the $X$ axis.
}
\label{fig:partial_prediction}
\end{figure*}

\subsection{Early Prediction of Contentiousness}
To predict contentiousness on partial conversations, we evaluated the previously trained models on increasing percentages of each conversation's comments. From a practical perspective, stakeholders such as moderators could be interested in solutions that predict contentiousness early with high performance \textit{i.e.} without seeing all the comments or seeing only a limited number of comments in addition to the titles and posts. 
To this end, we test both TF-IDF + All and BERT + All models by gradually increasing the proportion of conversation comments in the order they were posted, beginning at $10\%$ and ending at $100\%$ and increasing in $10\%$ intervals. The results of this experiment for each topic are shown in Figure~\ref{fig:partial_prediction}. 

Overall, we found that the inclusion of more comments and features increases the accuracy and F1 score significantly for TF-IDF + All. BERT + All also improves over time, but less substantially. TF-IDF + All approaches the performance of BERT + All at $100\%$ of comments, but BERT + All is much better at early stage prediction than TF-IDF + All. Specifically, models can obtain $88\%$ of peak performance for accuracy after the first $20\%$ of comments have been included, consistent across all three topics. The new features that we explore in this work provide significant performance benefits over the baseline, while remaining explain-able, both in full-prediction and partial prediction contexts.
\section{Explanation of Contentious Case-Studies} 

\begin{figure*}[!ht]
  \includegraphics[width=\textwidth, scale=0.8]{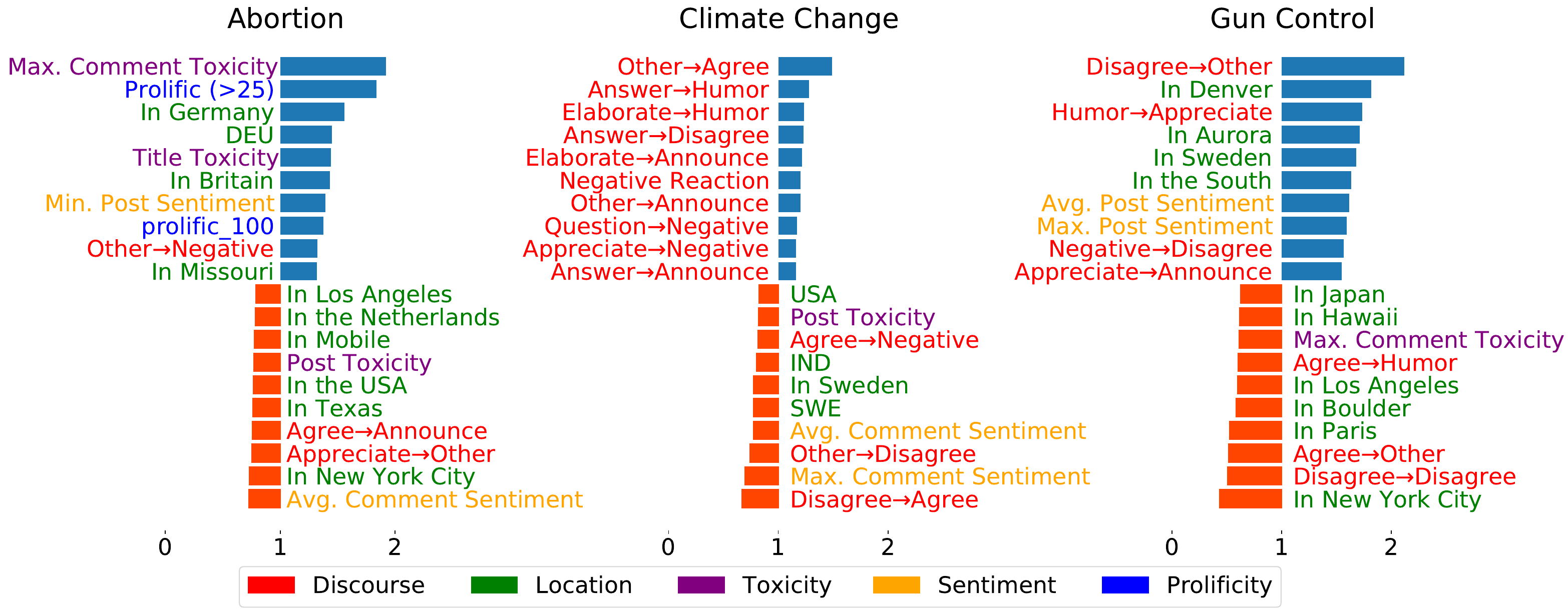}
  \caption{Odds ratios of the top 20 categories for contentious and non-contentious conversations across ``abortion'', ``climate change'', and ``gun control''. Feature names are colored based on their category. Prolificity, negative discourse acts, and title toxicity are generally associated with contentiousness across all three topics. Location, other discourse acts, and sentiment vary across topics. The feature importance across topics indicates a divide between contextual features and generally predictive ones.
  }
  \label{fig:odds_ratios}
\end{figure*}

This section investigates RQ2 on how different linguistic and conversational characteristics of conversations relate to their contentiousness and how these differ between different contentious topics. 
To do so, we utilize the regression coefficients from the aforementioned explainable classifier (TF-IDF + All), and further convert them into odds ratios for ease of explainability. 
We visualize the top 10 predictors with high odds ratios as well as the bottom 10 predictors with low odds ratios for each topic in Figure~\ref{fig:odds_ratios}. 

\subsubsection{Negative Discourse Acts Matter}
The presence of certain discourse acts, in particular \textit{disagreement}, \textit{negative reaction}, and \textit{question} are indicative of contentiousness across all three topics, with odds ratios greater than $1$ for each. \textit{Disagreement} and \textit{negative reaction} are discourse acts with a primarily negative connotation, so their appearance here is to be expected. \textit{Question}, however, is less so. Its appearance here indicates that the communication between the various sides of a divisive topic highly contributes to conversation contentiousness. Since \textit{answer} and \textit{elaboration} are significantly lower in terms of odds ratio, this would imply that communication is either one-sided or incomplete in these contentious conversations.

The discourse act bigrams that include \textit{disagreement}, \textit{negative reaction}, and \textit{question} are also heavily represented in contentious conversations, with the vast majority of them having odds ratios greater than 1. 
This is even the case when paired with positive discourse acts, such as \textit{agreement} \textit{elaboration}, and \textit{appreciation}, the cause of which may be that negative discourse acts are causing an effect similar to that of the negativity bias~\cite{pratto1991automatic,soroka2019cross}, a psychological phenomenon where negative events have greater psychological effects than positive ones, in the readers of these comments.

The positive discourse acts \textit{elaboration}, \textit{agreement}, and \textit{appreciation}, however, have significantly lower average odds ratios, $1.002$, $1.017$, and $0.984$, respectively. Bigrams containing these positive discourse acts similarly have lower odds ratios.. Each of these discourse acts indicates a greater level of inter-group cooperation and communication within these divisive topics. The least controversial act \textit{announcement} can be characterized generally as comments which are repositories of information, usually laid out in list form. These tend to be highly upvoted comments which guide the course of the conversation.

\subsubsection{Location Varies Across Topics}  
Location is, as shown by Figure~\ref{fig:odds_ratios}, variable in terms of both which specific locations are represented and the importance of them across topics. Two classes of location can be seen in the features for abortion: locations which are generally more accepting of abortion and locations which are generally less accepting of abortion. We see examples of the first category in \textit{in Germany} (odds ratio of $0.993$), and \textit{in Europe} ($1.005$)
and examples of the second category in \textit{in Georgia} ($1.109$) and \textit{in North Texas} ($1.076$)\footnote{https://www.pewforum.org/religious-landscape-study/compare/views-about-abortion/by/state/}. The posts of these users, however, have very little to do with their place of residence. This may indicate that a wide spread of people with complex views on abortion or on a diminished importance of particular locations. We also notice a third category of locations, which are places which are heavily featured in news related to abortion, with \textit{Seattle}, \textit{San Diego}, and \textit{New York} having odds ratios of $1.202$, $1.203$, and $1.118$, respectively.

Climate change may present a more political view of location. California ($1.212$) and New York ($1.184$) both had high values for the topic of climate change, and are both heavily featured in national news about climate change as well as politics concerning climate change.
Both of these states are primarily represented by the United States' Democratic Party whose constituents 
who are, on average, significantly more concerned about climate change on average than those of the United States' Republican party.\footnote{https://www.pewresearch.org/science/2015/07/01/chapter-2-climate-change-and-energy-issues/}
Gun control, however, presents a very specific view of location and contentiousness, with locations generally associated with gun violence tragedies being highly predictive of contentiousness.
For instance, Aurora, Colorado (which has an odds ratio of $1.196$) where, in 2012, a shooting was carried out which killed 12 and injured 70.\footnote{https://en.wikipedia.org/wiki/2012\_Aurora,\_Colorado\_shooting} Many users that we have identified as being from the Denver area, very near to Aurora, have participated in contentious conversations specifically about the shooting. 
Humans tend to react to events spatially close to them more concretely and emotionally\cite{fujita2006spatial} which is a likely cause, along with the particularly volatile subject matter, for these conversations to become contentious.

The examination of locations in contentious conversations shows that location is highly variable across topics as well as within them. For abortion, location may only play a broader role in contentiousness prediction. In climate change, locations appear to be related to political affiliation. In gun control, locations may be extremely specific and related to similarly specific events. As such, detecting contentiousness in conversations concerning issues involving very specific, spatially close locations can be viewed very differently from broader issues involving more spatial separation. 

\subsubsection{Prolificity and Account Age Are Inversely Correlated with Contentiousness}
Prolificity and account age are two related user factors which relate to contentious conversations in very different ways. Prolificity tends to correlate with contentious conversations across all topics. This would indicate that users that are, to a certain extent, prolific within a given community are more likely to participate in potentially difficult conversations due to feeling more comfortable engaging with the community on these issues. For instance, conversations featuring high numbers of prolific users in the topic of abortion overwhelmingly featured religion, a topic which many are uncomfortable talking about in unfamiliar spaces.
 Non-contentious conversations still involved religion, but not nearly to the same extent. Account age, on the other hand, is correlated with a lack of contentiousness.
\subsubsection{Gender Predicts Abortion}
Gender has an extremely variable impact on contentiousness depending heavily on whether or not gender is an inherent issue for the contentious topic being discussed. 
This is immediately evident in our analysis: the presence of specific genders is only appreciably different when looking at abortion, with male having an odds ratio of $1.128$ compared to the female value of $1.061$. 
The fact that gender can be so influential in the topic of abortion as to raise the odds ratio of male so high is a testament to the gender differences that can appear in contentious conversations \cite{herring2020dominates} and that abortion is a topic that particularly concerns women more than men. Those studying conversations using our method in a particular community which may deal with female-centric issues more frequently than usual should pay more attention to gender when determining contentiousness. Notably, these results suggest that the differences in which female and male users engage with contentious discussion found in \citeauthor{herring2020dominates} are highly variable with topic. 
\subsubsection{Toxicity Equals Contentiousness?}
High levels of toxicity in post titles are strong predictors of contentiousness, as demonstrated by odds ratios that are greater than $1.1$ in all topics. It might be that some users choose contentious or provocative titles to attract more users to participate in conversations; similar trends have been found on Twitter~\citep{reis2015breaking}, where the sentiment of the news headline is strongly related to the popularity of the news.
As demonstrated in Figure~\ref{fig:odds_ratios}, however, the toxicity of other parts of conversations are weighted differently between topics. For instance, \textit{max. comment toxicity} contributes to contentiousness across all three topics while \textit{avg. comment toxicity} only contributes to contentiousness in abortion. This might be because different topics have their own characteristics, with people involved in discussions about climate change being more likely to use milder language in comments, with an average toxicity level of $0.19$. On the other hand, participants in abortion and gun control conversations are more likely to use more toxic language in comments, with an average toxicity level of $0.26$ and $0.25$, respectively. 
\textit{Selftext toxicity} 
is relatively less predictive, with odds ratios of less than $1$ among all topics. This might be because posts usually serve a pragmatic purpose, either as expansions or explanations of their title in order to provide more information about the poster's intent and context for discussions. It may also be that the content of the title colours commenters' opinions more strongly than the selftext, following the human preference for first impressions.

\subsubsection{LIWC Offers Contextual View of Language Choice on Contentiousness}

In general, LIWC features offer a comprehensive view of users' linguistic styles and word choices and their influences on controversy. For simplicity, here we discuss two consistently informative LIWC features: \textit{you} and \textit{friends}. 
Here the linguistic category \textit{you} contains second-person pronouns such as ``you'', and  ``yours''. \textit{you} is positively associated with contentiousness across all topics 
which might communicate to us that a repetitive utilization of \textit{you} words 
increases social distance between conversation participants, and relatively related to relationship quality \citep{simmons2008hostile}.
Such situations can be observed in our cases. 
The impact of the category \textit{friends} is more variable, with climate change viewing it the most indicative of contentiousness ($1.13$) and abortion viewing it the least ($0.91$). 
This might indicate that higher mentions of \textit{friend} words in abortion make them harder to serve as a distinction signal in contentiousness prediction, as abortion involves large amount of descriptions about relationships, for which people need to refer to related words related to friends. In both contentious and non-contentious conversations, we can easily find such discussions.
Conversely, we found that in climate change \textit{friends} tended to be associated with sarcasm, or negative experiences with friends. In general, these results are highly contextual with regards to target and community.

\section{Discussion and Conclusion}
This work investigated contentious discussions via case studies on three highly divisive topics: abortion, climate change, and gun control. Specifically, we operationalized a set of conversation characteristics ---
discourse acts, toxicity, sentiment, LIWC, text representation, and user factors --- and demonstrated that 
these features contributed to both reasonable predictive performances 
and explainability in understanding contentious in conversations.
In doing so we discovered that different linguistic and social features have different relationships to contentiousness and the topics themselves. Certain features --- toxicity, prolificity, and discourse acts --- remained stable predictors of contentiousness regardless of the topic. Other features --- location, gender, sentiment, and LIWC --- demonstrated highly contextual information and differences in the ways in which users engage with these contentious topics on Reddit.
The findings suggest that contentiousness in conversations is subtle and that explainable models are necessary to identify these nuances for both moderators, users, and researchers.   

\subsection{Implications}
Our work has both theoretical and empirical guidelines to assist online media in defining, measuring, understanding, and even intervening with contentious content on social media platforms.  
First, our research provides conversational characteristics of contentious conversations which enable deep, contextual insights into the nature of these conversations.
Second, our proposed generic conversation characteristics can be
easily applied to build predictive models for understanding other similarly divisive topics beyond the three studied here, since they are not tailored to any specific data.

Practically, this work could motivate the design of tracking/monitoring systems that can visualize possibly contentious conversations in any specific groups to assist community moderators with content moderation. In addition to warning moderators early on in a post's lifespan whether or not it is of concern, this work has the ability to explain to moderators (and users) why a post is considered contentious and has attracted increased attention. Since the explanation of moderator action is a large contributor to user satisfaction~\cite{suspectpost}, the explanation portion of our research is key for ensuring quality moderation strategies can be derived from our work. It also sheds light on designing better features to inform new community members on what aspects of the conversations might cause problematic experiences, allowing for increased retention of newer users who are not exposed to contentious content which they may not wish to engage with early on in their time on the subreddit. Since there is some amount of subject-matter familiarity required to properly explain the models we have proposed here, moderators which are very familiar with the content that is posted to their communities naturally fit this role. Our work will also help design novel tutorials to educate users on how to engage in discussions, especially in divisive topics, in a more civil and constructive manner. Particularly, these tutorials can be tailored to individual topics with the use of this work and subject-matter experts.

\subsection{Generalization}
Although our work focused only on three topics: abortion, climate change, and gun control, many of our features and characteristics of these divisive topics can generalize well to other topics. LIWC, toxicity, and sentiment have been shown to be powerful predictors of contentiousness from previous work\cite{hessel2019something,mejova2014controversy,dori2013detecting}. We expect the patterns that we have observed regarding user factors, excluding location, to generalize well. 
For gender, the strong correlation of gender with contentiousness in the topic of abortion suggests that topics which relate to gendered issues will hold a similar relationship to gender, while non-gendered issues will continue to display little to no relationship with gender. We do not expect that the same locations should emerge in a generalized version of this work, although that locations which are relevant to the issue being discussed will continue to appear. In a generalized version of this work, less specific measures of location diversity (e.g., entropy) may perform better. We also hypothesize that discourse acts will hold generally for contentious conversations, and leave confirmation of that hypothesis to future work.  

\subsection{Limitations}
Our work also has a few limitations.
First, the distant-supervision method that we use to label the contentiousness in posts relies on upvote ratio, which may not be a perfectly accurate analogue for contentiousness, though this has been adopted and validated by prior work \cite{hessel2019something}. The Rediquette  (Reddit's rules of conduct) specifically states that users are not to ``downvote an otherwise acceptable post because you do not personally like it.'' Although only a subset of users follow these recommendations, this indicates that the downvote button is mainly reserved for posts that break rules or do not belong on a particular subreddit. We partially circumvent this limitation by controlling for topic. 
Future work can conduct further annotations to validate the correlations between human judgement and our contentiousness measures. 
Another important limitation of this work is its correlational
nature.
We mainly characterize which features of conversations are correlated with contentiousness but not what causes that relationship. 
However,
without true random-assignment experimentation we cannot
assume that the relationships described reflect any degrees of
causation.

\subsection{Ethics Statement}
This research study has been approved by the Institutional Review Board (IRB) at the researchers' institution. 
In this work, we leverage no information that was not publicly available at the time of data collection, leveraging the public post histories of the users who participated in the conversations which we studied. As such, user private information is not disclosed, and none of the posts which were used to compute the gender and location features have been saved or need to be saved during the course of this computation.

\bibliography{main}
\end{document}